\documentclass{article}

\usepackage{PRIMEarxiv}

\usepackage[utf8]{inputenc} 
\usepackage[T1]{fontenc}    
\usepackage{hyperref}       
\usepackage{url}            
\usepackage{booktabs}       
\usepackage{amsfonts}       
\usepackage{nicefrac}       
\usepackage{microtype}      
\usepackage{lipsum}
\usepackage{fancyhdr}       
\usepackage{graphicx}       
\graphicspath{{media/}}     

\usepackage{amsmath,amssymb,amsfonts}
\usepackage{algorithmic}
\usepackage{algorithm}

\usepackage{textcomp}
\usepackage{booktabs} 
\usepackage{multirow}

\usepackage{float}
\usepackage{cleveref}

\usepackage{threeparttable}
\usepackage{array}

\crefname{figure}{fig.}{figures}
\Crefname{figure}{Fig.}{Figures}

\crefname{table}{table}{tables}
\Crefname{table}{Table}{Tables}

\crefname{algorithm}{algorithm}{algorithms}
\Crefname{algorithm}{Algorithm}{Algorithms}

\title{CSubBT: A Self-Adjusting Execution Framework for Mobile Manipulation System
\thanks{\textit{This work has been submitted to the IEEE for possible publication. Copyright may be transferred without notice, after which this version may no longer be accessible.}} 
}

\author{
  Huihui Guo, Huizhang Luo, Huilong Pi, Mingxing Duan, Kenli Li\\
  Hunan University \\
  Changsha\\
  \texttt{\{ghh1991@hnu.edu.cn, lkl@hnu.edu.cn\}} \\
   \And
  Chubo Liu \\
  Hunan University \\
  Changsha\\
  \texttt{liuchubo@hnu.edu.cn} \\
}

\begin{document}
\maketitle

\begin{abstract}
        With the advancements in modern intelligent technologies, mobile robots equipped with manipulators are increasingly operating in unstructured environments.
        These robots can plan sequences of actions for long-horizon tasks based on perceived information.
        However, in practice, the planned actions often fail due to discrepancies between the perceptual information used for planning and the actual conditions.
        In this paper, we introduce the {\itshape Conditional Subtree} (CSubBT), a general self-adjusting execution framework for mobile manipulation tasks based on Behavior Trees (BTs).
        CSubBT decomposes symbolic action into sub-actions and uses BTs to control their execution, addressing any potential anomalies during the process.
        CSubBT treats common anomalies as constraint non-satisfaction problems and continuously guides the robot in performing tasks by sampling new action parameters in the constraint space when anomalies are detected.
        We demonstrate the robustness of our framework through extensive manipulation experiments on different platforms, both in simulation and real-world settings.
\end{abstract}


\section{INTRODUCTION}

With the development of artificial intelligence, robots require the capability to perform long-horizon tasks to address the complex and varied demands of unstructured environments.
Integrating task planning algorithms on mobile platforms equipped with basic sensing, autonomous navigation, and manipulation capabilities has become a standard configuration.
Task and Motion Planning (TAMP) is a recent method \cite{b1,b2} that generates sequences of symbolic actions for mobile manipulation tasks.
The symbolic actions, typically defined by a domain-specific planning language (e.g., PDDL) \cite{b2.1}, 
encapsulate essential geometric details, such as the pose of an object to be grasped during a pick operation, thereby effectively guiding the robot in executing the task.

Nevertheless, the geometric information available to the robot before planning may not always be current or accurate.
Additionally, sensor deviations and disturbances are common during task execution, especially when multiple points of robot movement are involved.
Traditional TAMP approaches have primarily focused on solving planning problems under ideal conditions or without the necessity to alter the robot's position \cite{b2}.
Many executors of these approaches execute commands directly without sufficient robustness.
Most approaches in the anomaly handling literature focus on preventing potential failures by
(1) manually developing strategies to minimize the likelihood of such failures occuring,
(2) relying on the replanning capabilities of high-level planners to address unforeseen anomalies, and
(3) employing proficient actions learned from diverse experiences, including those involving failure \cite{b14, b8, b14.2}.

\begin{figure}[t]
        \centerline{\includegraphics[width=\columnwidth]{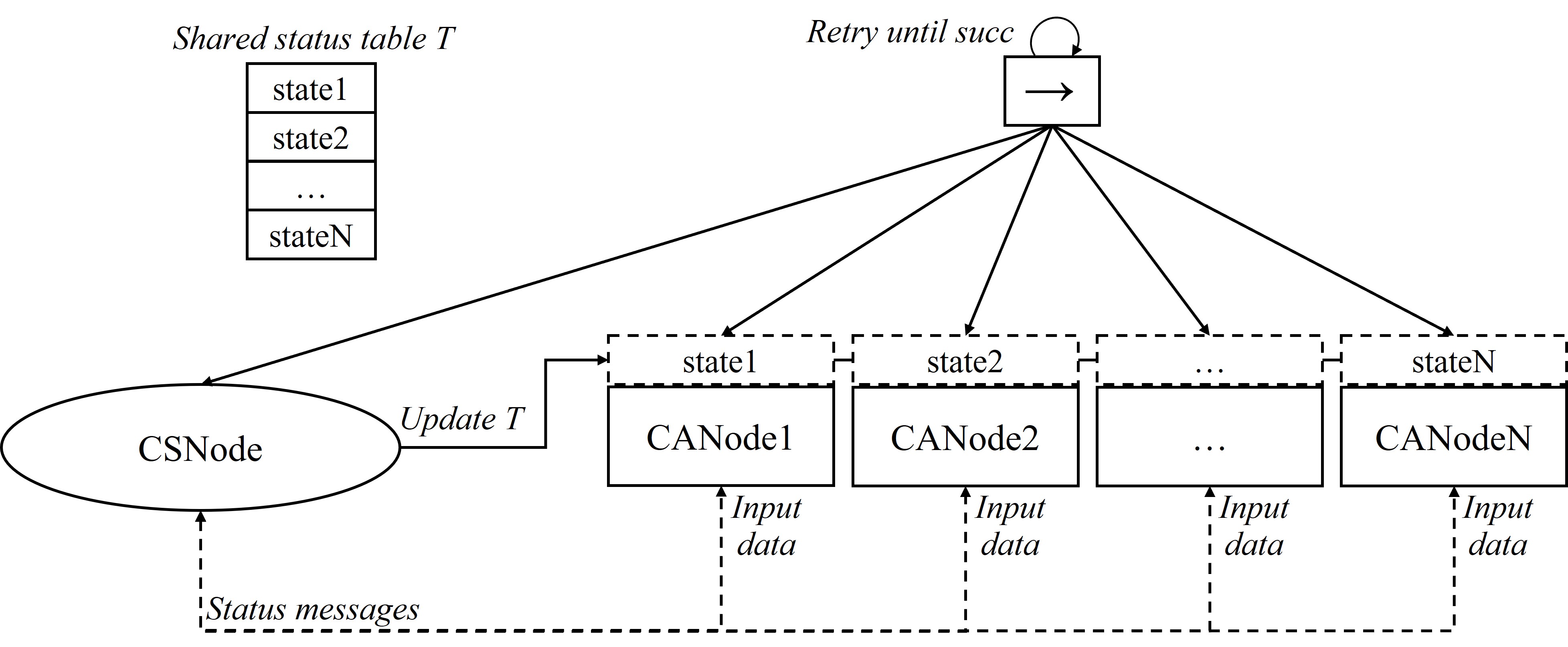}}
        \caption{A general structure of CSubBT.
                The execution framework is based on BTs.
                CSNode collects and handles error information uniformly,
                then transmits adjustment parameters and the boolean state to each CANode through the {\itshape blackboard} to address errors.
                More details are in Section \ref{section: CS Node and CA Node}.}
        \label{fig_framework}
\end{figure}

However, developing anomaly handling strategies are often difficult to generalize across different applications and demand extensive manual effort.
In long-horizon tasks, replanning is typically time-consuming \cite{b3, b3.1}, and it is not cost-effective to replan the entire task due to common anomalies (e.g., sensor errors).
While the learning-based methods require substantial training data,
it remains crucial to design a robust and general execution framework for long-horizon tasks to improve the overall success rate of task completion.
In \cite{b5, b7}, the planner samples the geometric parameters of actions from the constraint space and forms feasible motion instructions.
However, in the actual execution process, common sensor deviations or dynamic disturbances can cause these motion commands to fail, resulting in action failures.
But if these samplers are reused to adjust motion parameters during execution, giving the executor the ability to dynamically adjust parameters,
it can make the execution process of actions more robust and avoid high-level replanning to handle anomalies caused by sensor errors.

In this paper, we introduce CSubBT, a robust, self-adjusting execution framework (\Cref{fig_framework}) based on BTs for implementing mobile manipulation tasks.
CSubBT is a modular subtree comprising a {\itshape Conditional Switch Node} (CSNode) and several {\itshape Conditional Action Nodes} (CANodes).
A CANode is an {\itshape Action} node in BTs with an additional Boolean state identifier that determines whether it should execute. 
The CANodes in CSubBT execute sequentially based on their state identifiers.
A CSNode is a {\itshape Condition} node in BTs that 
contains Conditional Samplers \cite{b5} within the motion constraint space and is used to manage the state identifiers of CANodes in CSubBT.
The execution of a general CSubBT represents the execution of a symbolic action.
A CANode represents an atomic action, which is a finer-grained sub-action derived from a symbolic action.
For example, the symbolic action {\itshape OpenDoor} requires the robot to first adjust its base pose, 
then adjust the robotic arm to the appropriate pre-approach pose, and finally to pull the door handle.
We model a symbolic action or a newly composed symbolic action as a factored transition system \cite{b5}, such as the symbolic actions {\itshape OpenDoor} and {\itshape Move-and-Pick}.
CSubBT leverages the approach described in \cite{b5} to analyze the constraint relationships between the atomic actions of a symbolic action, 
enabling exploration of the constraint space without requiring manually defined anomaly handling strategies or high-level replanning.
When an action execution fails under the initial parameters, CSubBT resamples new parameters within the constraint space to retry the action. 

The primary contribution of this paper is the direct implementation of symbolic actions from a high-level planner into CSubBTs for robust execution. 
We formulate the execution of CSubBT as a problem of finding feasible solutions within the constraint space. 
By integrating parametric sampling and repeated execution of factored atomic actions, 
CSubBT effectively explores the constraint space, enhancing its overall success rate. 
Experimental results demonstrate the robustness of CSubBT and its improved reactivity in handling anomalies.

\section{Related Works}

In recent years, the AI planning community has been exploring the use of intelligent robots to plan and execute long-horizon tasks in unstructured environments.
Nicola Castaman et al. \cite{b8} proposed an online planner that integrates a low-level motion planner with a task planner in a receding horizon mode.
This approach executes only the first action in the plan, allowing the planner to respond to dynamic changes as they occur.
However, this method assumes that the sensors detecting the environment's state are operating ideally and capturing information in real-time.
This assumption is challenging to meet when robots are performing tasks, particularly in unstructured environments.
Without sufficient attempts, true anomalies are difficult to identify directly.
Furthermore, if the geometric information provided to the planner is not sufficiently accurate or if there is some bias during execution,
it is more practical to utilize low-level motion planners to address the problem during the actual execution process.

As early as 2010, Weser et al. \cite{b9} highlighted the necessity for mobile autonomous robots to have closed sensor control loops and pre-planned solutions for complex tasks.
They integrated the Hierarchical Task Network (HTN) planner with physical perception and execution of atomic actions, avoiding the closed-world assumption (CWA).
This approach offers various methods for managing fixed exceptions, such as invalid world representations, device failures, and execution errors.
The HTN planner decomposes complex tasks into sub-actions based on task domain files.
However, since the HTN planner relies on task searching, it requires relevant domain knowledge for replanning when anomalies occur.
In contrast, our paper presents a reactive executor for high-level planners, whereas HTN is difficult to decouple as a standalone executor for the planner.

D. C. Conner et al. \cite{b10} utilized Finite State Machines (FSMs) as behavior controllers for action execution in the Flexible Manipulation system.
FSMs handle predictable, repeatable motions based on sensor inputs.
However, FSMs become complicated when handling exceptions, especially with numerous states,
making it difficult to directly use FSM as a robust executor in unstructured environments.
As a replacement for FSMs, Behavior Trees (BTs) offer improved modularity and reactivity \cite{b11.1}.
In \cite{b13}, the system combines the Stack-of-Tasks control strategy, serving as high-level decision logic, with BTs.
Although their method completes tasks robustly and reacts to unexpected changes,
it is limited by the need to design a complex BT logic structure when dealing with tasks with multiple constraints, such as mobile manipulation tasks.

Ruichao Wu et al. \cite{b14} proposed a framework for error recovery in robotic manipulation systems
using an external Bayesian Network module to infer the causality of errors and generate recovery actions for BTs.
However, their approach relies on manual specification of exception-handling policies and lacks analysis and exploration of the constraint space.
In \cite{b14.1}, a new robot control platform, SKiROS2,
is proposed for automated task planning and reactive execution, combining BTs and FSMs to constitute a reactive execution system.
Similar to our work, they define primitive and compound skills to satisfy different task requirements.
Nevertheless, their primary objective is to construct a comprehensive system, without in-depth analysis of the causes of action execution failures.

Mitrevski et al. \cite{b14.2} present an execution model (XM) and conduct a comprehensive analysis of the underlying causes of execution failures.
XM is a learning-based hybrid representation for parameterized actions.
The relevant action parameters are obtained through model prediction, whereas our method samples adjustable parameters from the relevant constraint space.

In this work, we propose CSubBT, an execution framework based on BTs. 
CSubBT achieves robust execution by decomposing high-level symbolic actions into a sequence of atomic actions and controlling their executions. 
Following the approach outlined in \cite{b5}, CSubBT systematically analyzes the constraint relationships between these atomic actions, 
framing common execution anomalies as constraint satisfaction problems.
In contrast to their recent work on partially observable environments \cite{b7.1}, CSubBT's exception-handling capabilities are confined to the constraint space of its actions.
To handle such anomalies, CSubBT dynamically adjusts the geometric parameters of the atomic actions by sampling within the constraint space until either the action succeeds or all viable parameter configurations are exhausted.

\section{Background}\label{background}

\subsection{Mobile Manipulation}

The nature of mobile manipulation tasks involves a combination of navigation and manipulation subtasks.
This process is augmented with real-time sensory information to ensure successful implementation.
Autonomous navigation and robotic manipulation in static environments are well-established areas of research \cite{b14.5,b15}.
In this work, we decompose and reorganize mobile manipulation tasks based on the concepts and strategies proposed in \cite{b5} and \cite{b17}.

\subsubsection{Factored Transition System}

The mobile manipulation problem can be effectively decomposed into comprehensive subproblems with phased multiple constraints.
Ideally, the problem can be modeled as a factored transition system $S = \left\langle \mathcal{X},\mathcal{U},\mathcal{T}\right\rangle $,
where $\mathcal{X}$ is a set of states, $\mathcal{U}$ is the control space and
$\mathcal{T}\subseteq \mathcal{X}\times \mathcal{U} \times \mathcal{X}$ is the transition relation.
Given a set of goal states $\mathcal{X}^*\subseteq \mathcal{X}$ and an initial state $\bar{x} ^0\subseteq \mathcal{X}$,
a sequence of control inputs ($\bar{u} ^1$, $\bar{u} ^2$, ...,$\bar{u} ^n$) constitutes the plan for the problem.
The system can then be factored as transitions $(\bar{x} ^{i-1},\bar{u} ^i,\bar{x} ^i)\in \mathcal{T}$ for $i\in \{1,...,n\}$ ensuring that $\bar{x} ^n\in \mathcal{X}^*$.

The problem of mobile manipulation is a hybrid constraint satisfaction problem in a high-dimensional space.
The factored transition system can expose the constraints $(C_1, C_2, ..., C_n)$ of subproblems.
Moreover, most of these constraints can be evaluated as functional descriptions with equality form in the configuration space of the robot.
Equality constraints can significantly simplify the space of transition parameters.
The factored transition relation is a union of several transition components, $\mathcal{T} = \cup ^\alpha _{a=1}T_a$.
A {\itshape transition component} $T_a$ often lies at the intersection of the constraint manifold with a subset of the state parameters.
The constrained manifold is a set of variables that satisfy certain constraints, typically manifested as a subset of high-dimensional space.
The intersection of $T_a$ can be described as a {\itshape clause} of $\beta $ constraints $\mathcal{C}_a = \{C_1, ..., C_\beta \}$,
where $T_a = \bigwedge ^\beta _{b=1}C^a_b$.

The plan parameter space for the mobile manipulation problem is an alternating sequence of states and controls
$(\bar{x} ^{0}, \bar{u} ^1, \bar{x} ^{1},\dots,\bar{u} ^*,\bar{x} ^{*})\in \bar{\mathcal{X}}\times (\bar{\mathcal{U}}  \times\bar{\mathcal{X}} )^*  $.
The solutions in the parameter space must satisfy a plan-wide
{\itshape clause} of constraints $\mathcal{C}_{\vec{a}} = \mathcal{C}_0\cap \mathcal{C}_1 \cap \dots \cap \mathcal{C}_*$,
where $\mathcal{C}_{\vec{a}}$ is referred to as a {\itshape constraint network}.
This constraint network is a bipartite graph between constraints and parameters \cite{b18,b19}.
The {\itshape constraint network} enables a transparent and unambiguous representation of the parameters and constraints inherent to the transition system throughout the planning process.

\subsubsection{Conditional Samplers}

Implicit representations of constraints are widely used in the research area of sampling-based motion planning, such as collision checking.
Traditional samplers either sample values deterministically or non-deterministically.
By analyzing the {\itshape constraint network} and explicitly characterizing the intersection manifolds,
we can reduce the dimensionality of the associated constraint space.
For example, the goal state $x^1_b$ for the {\itshape Move} action can be sampled from an uncountably infinite set.
However, if the next {\itshape Pick} action requires reaching an object on the desk, we only need to generate the object location from the neighborhood of the desk.

A {\itshape conditional sampler} $\psi = \left\langle I, O, \mathcal{C} , f\right\rangle $ is represented by a function $f(\vec{x}_I )=(x^1_O,x^2_O,...)$,
where $\vec{x}_I$ is the set of input parameters, $x_O$ is the sequence of sampled values and $\mathcal{C}$ represents the related constraints.
Relatively, any $\psi$ that has no inputs works as an {\itshape unconditional sampler}.
A transition system often has a sequence of conditional samplers $\vec{\psi } = (\psi _1,...,\psi _n)$.
The set of values generated by $\vec{\psi }$ is given by:\par
{\small
        \begin{equation}
                F(\vec{\psi } ) = \{\vec{x} |\vec{x}_{O_1}\in f_1(\vec{x}_{I_1} ), \vec{x}_{O_2}\in f_2(\vec{x}_{I_2} ),...,\vec{x}_{O_n}\in f_n(\vec{x}_{I_n} ) \}.
        \end{equation}
}%

Generally, a transition system is robustly feasible if the constraints in the finite sequence of transition components are satisfiable with some redundancy.
A set of conditional samplers is sufficient for a robustly feasible transition system
if there exists a sequence of samplers $\vec{\psi }$ such that $F(\vec{\psi } )\cap  \vec{\mathcal{C} }\neq \emptyset  $.

\subsubsection{Grasp Planner}

A hierarchical grasp planner connects states from the start and end interfaces of each stage, combining them into a feasible trajectory solution.
In the generation stage, intermediate states can be provided by the planner's grasp generator, which acts as a conditional sampler.
The grasp generator first identifies an exhaustive set of possible grasp candidates based on specified conditions, then evaluates their quality to prioritize them.
Issues in connecting stages can be addressed using existing motion planning frameworks, such as MoveIt!.
These frameworks generate interpolated and smoothed Cartesian paths based on current and target states.
Additionally, explicitly factoring manipulation tasks into maintenance-friendly stages and world states aids in anomaly analysis \cite{b17},
thereby enhancing the robustness of task execution.

\subsection{Behavior Trees}

In recent years, BTs have been increasingly utilized in intelligent robotic systems, originating from their use in gaming applications \cite{b4}.
BTs are composed of control flow nodes ({\itshape Sequence}, {\itshape Fallback}, {\itshape Parallel}, {\itshape Decorator}) 
and execution leaf nodes ({\itshape Action}, {\itshape Condition}).
The symbols and operational mechanisms of BT nodes are illustrated in \Cref{BT nodes}.
The control flow nodes route the {\itshape tick} signals from their left children to the right,
returning either {\itshape Failure}, {\itshape Running} or {\itshape Success} based on the execution outcome of the child nodes.
Furthermore, a common information pipeline, known as the {\itshape blackboard}, is used to pass parameters between BT nodes.

\begin{table}[h]
        \caption{BASIC NODE TYPES OF A BT}
        \label{BT nodes}
        \centering
        \resizebox{\linewidth}{!}{
                \begin{threeparttable}
                        \begin{tabular}{ccccc}
                                \toprule[2pt] 
                                \multirow{2}*{Node Types} & \multirow{2}*{Symbols}                                      & \multicolumn{3}{c}{Return Status}                                                                                                            \\
                                \cline{3-5} 
                                                          &                                                             & Success                                                & Running                                  & Failure                                  \\
                                \cline{1-5} 
                                Sequence                  & \includegraphics[width=0.015\textwidth]{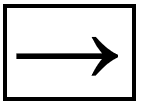}  & Only if all children return Success                    & The current ticked child returns Running & The current ticked child returns Failure \\
                                Fallback                  & \includegraphics[width=0.015\textwidth]{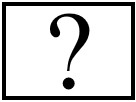}  & The current ticked child returns Success               & The current ticked child returns Running & Only if all children return Failure      \\
                                Parallel\tnote{*}         & \includegraphics[width=0.015\textwidth]{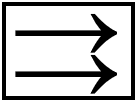}  & If M children return Success                           & otherwise                                & If N - M + 1 children return Failure     \\
                                Decorator                 & \includegraphics[width=0.015\textwidth]{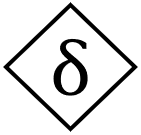} & \multicolumn{3}{c}{varies based on user-defined rules}                                                                                       \\
                                Action                    & \includegraphics[width=0.015\textwidth]{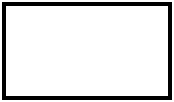}    & If Action completes successfully                       & During execution                         & If Action fails                          \\
                                Condition                 & \includegraphics[width=0.015\textwidth]{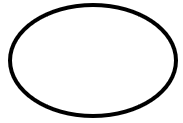} & If Condition is satisfied                              & Never                                    & If Condition is not satisfied            \\
                                \bottomrule[2pt] 
                        \end{tabular}

                        \begin{tablenotes}    
                                \footnotesize               
                                \item[*] N is the number of children and M is a user-defined threshold that is not greater than N.
                        \end{tablenotes}            
                \end{threeparttable}       
        }
\end{table}

BTs provide a graphical modeling framework that decomposes complex tasks into modular subtasks.
The Postcondition-Precondition-Action (PPA) and Backchaining are the typical design principles of BTs \cite{b4}.
Based on these principles, 
the combination of control flow nodes and execution leaf nodes facilitates the construction of complex, reactive execution systems.
However, selecting the appropriate level of granularity is crucial due to the modular nature of BTs. 
Specifically, it is essential to balance the tree's structure, 
ensuring that it is neither excessively complex nor too compact, as either extreme could compromise its reactivity.

\section{Proposed Approach}\label{approach}

In mobile manipulation applications, a feasible plan can be described as $\pi =[a_1(\bar{x}_1),\dots, a_n(\bar{x}_n)]$, where $a_n$ is the general action defined with symbolic languages, such as PDDL,
and $\bar{x}_n$ is a set of relevant parameters.
From the perspective of engineering implementation, most of the actions are factorable,
for example, when executing the common symbolic actions {\itshape Pick}, {\itshape OpenDoor} and {\itshape CloseDrawer} in classic planning tasks,
the mobile manipulator has to follow a pattern of movement that begins with preparation, continues with observation and ends with operation.
Besides, if some actions can not be factored, they can often be combined with neighboring actions in the plan to form a new factorable action,
such as the {\itshape Move} action.
Therefore, the plan can be described as $\pi =[A_1(\bar{x}_1),\dots, A_m(\bar{x}_m)]$,
where $1 \leqslant m \leqslant n $.

The factorable action $A_m$ can be treated as a factored transition system, where the transition relation $\mathcal{T}_m = \cup ^\alpha _{a=1}T_a$ comprising transition components $T_a$.
Each transition component corresponds to an atomic action factored by $A_m$.
As mentioned in \Cref{background}, $T_a$ lies in the intersection of constraints manifold $\mathcal{C}_a$.
The relevant parameters $\bar{x}_m$ of $A_m$ can be extracted from the samplers $\psi_m$ of the manifold.
Since most of the transition systems possess a {\itshape constraint network} \cite{b5},
it graphically describes the process of parameter sampling.

Once the parameters in the action sequence are identified, we can implement them using the proposed execution framework.
To illustrate our methodology, we will detail the factorable action {\itshape Move-and-Pick} as an example.
The initial plan $\pi$ we assume is obtained from the high-level planner.

\subsection{The Move-and-Pick Process}\label{The Move-and-Pick Process}

Synthesizing the experience of most mobile manipulation processes, it is crucial to first localize the object's pose before attempting to grasp it.
Determining the appropriate grasping pose is essential for ensuring the stability and robustness of the manipulation process.

\begin{figure}[h]
        \centerline{\includegraphics[width=\columnwidth]{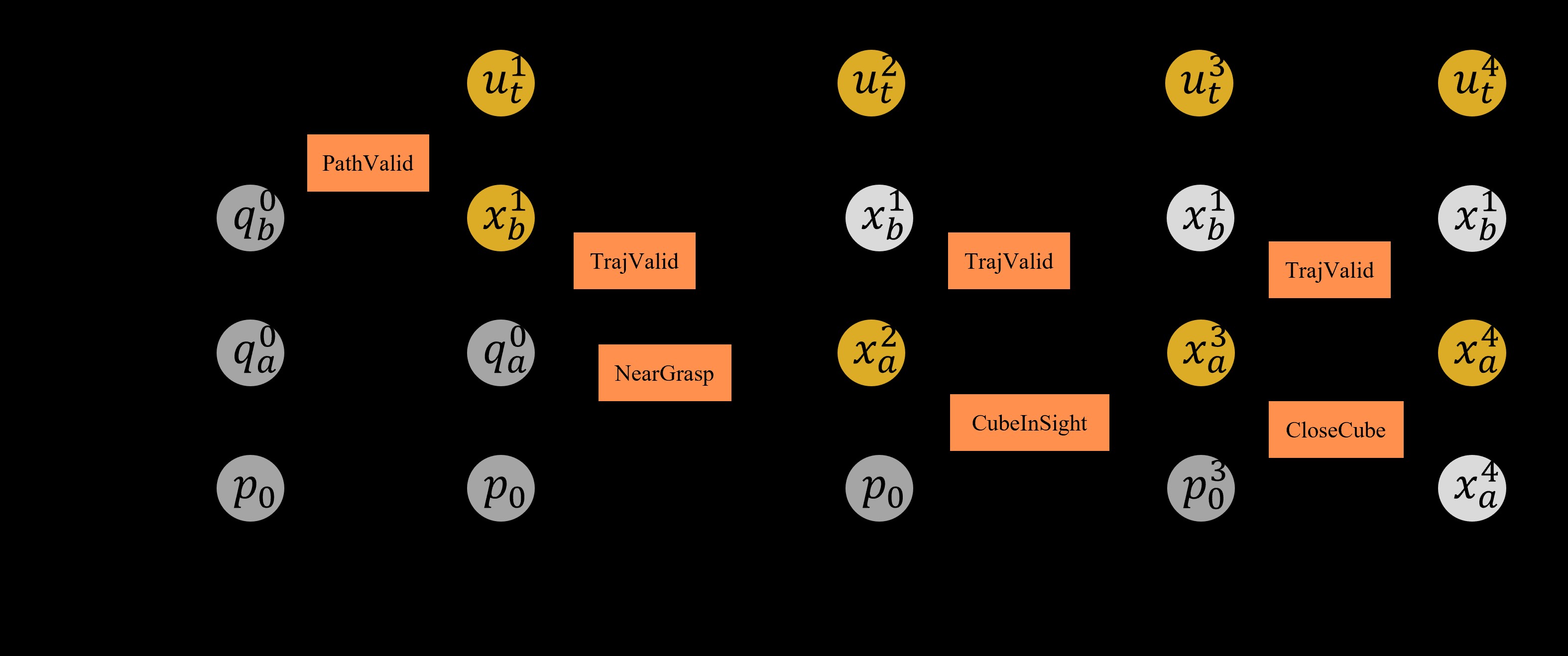}}
        \caption{The {\itshape constraint network} for the {\itshape Move-and-Pick} process.}
        \label{fig_constraint_network_eg}
\end{figure}

We combine the actions {\itshape Move} and {\itshape Pick} into a new factorable action called {\itshape Move-and-Pick}.
We then factor {\itshape Move-and-Pick} into four atomic actions: {\itshape Move}, {\itshape Pre-approach}, {\itshape Approach}, and {\itshape Grasp}.
The exact number of atomic actions to split into depends on the specific method of configuration.
The corresponding {\itshape constraint network} is illustrated in \Cref{fig_constraint_network_eg}.
Dark grey circles represent constant state parameters with equational constraints, yellow circles indicate free parameters,
and light grey circles denote two equal transient fixed parameters.
Orange rectangles indicate constraints.
$x_a$ and $x_b$ denote the arm states and base states of the robot, respectively.
The transition relation $\mathcal{T}$ has four {\itshape clauses},
and is represented as
\begin{center}
        $\mathcal{T} = \{\mathcal{C}_{Move}\cup \mathcal{C}_{Pre-approach}\cup \mathcal{C}_{Approach}\cup \mathcal{C}_{Grasp}\}$.
\end{center}
We assume that manipulation actions occur only when the base is stationary and that the visual information of the cube is provided by the camera in hand.
This is necessary because the calculation of the grasping object's pose depends on the relative position of the base.

The {\itshape Move} action provides the capability to change the position of the base.
The transition component of this action, $T_{Move} = {\mathcal{C}_{Move}}$, contains a single constraint {\itshape clause}:
\begin{center}
        $\mathcal{C}_{Move}=\{ PathValid\} $.
\end{center}
The action must satisfy the {\itshape PathValid} constraint during execution, which ensures a valid or collision-free path.
The tool center point (TCP) of the hand must be moved to the pre-approach pose to detect and approach the object,
the action is called {\itshape Pre-approach}, the constraints {\itshape clause} of transition $\mathcal{T}_{Pre\text{-}approach}$ is:
\begin{center}
        $\mathcal{C}_{Pre\text{-}approach}=\{ NearGrasp, TrajectoryValid\} $.
\end{center}
The path and arm trajectory are generated using the existing motion planning framework.

Once the object is detected, the {\itshape Approach} action moves the arm in Cartesian space to maintain a consistent vertical distance between the object and the TCP,
the related constraints {\itshape clause} of transition $\mathcal{T}_{Approach}$ is:
\begin{center}
        $\mathcal{C}_{Approach}=\{CubeInSight\}$.
\end{center}
Finally, the {\itshape Grasp} action completes the picking process, the constraints of $\mathcal{T}_{Grasp}$ is described as:
\begin{center}
        $\mathcal{C}_{Grasp}=\{ CloseCube,CubeInSight\}$.
\end{center}
Considering that the {\itshape CubeInSight} and {\itshape CloseCube} constraints depend on the perceptual algorithm,
we assume these constraints are satisfied before executing.
The grasp trajectories of atomic actions {\itshape Approach} and {\itshape Grasp} are assumed valid when the {\itshape CubeInSight} and {\itshape CloseCube} constraints are satisfied.
As a result, we only need to determine $x_a$ and $x_b$ before obtaining the pose of the cube.

The {\itshape Move} action involves samplers for the constraints $Var_q^b$ and {\itshape BaseMotion}.
$\psi_B$ is an unconditional sampler for $Var_q^b$, and $\psi_T$ is a conditional sampler for {\itshape BaseMotion}, where
\begin{center}
        $\psi_B = \langle(),(x_q^b),{Var_q^b}$,SAMPLE-CONF$\rangle$,
\end{center}
\begin{center}
        $\psi_T = \langle(x_0^b,x_q^b),(u_t^b),{BaseMotion}$,STRAIGHT-LINE$\rangle$.
\end{center}
Considering the {\itshape NearGrasp} constraint, the SAMPLE-CONF process samples $x_q^b$ near the desk where the cube is,
and the STRAIGHT-LINE process contains the generating process of $x_q^b$ and
calculates a collision-free path $u_t^b$ between the start state $x_0^b$ and the target state $x_q^b$.
Since the sampler $\psi_B$ requires information on the desk and robot size as parameter inputs, we have combined these two samplers into one conditional sampler:
\begin{center}
        $\psi_{Move}= \langle(x_0^b,x_q^b),(u_t^b),\{PathValid\}$,SAMPLE-MOVE$\rangle$.
\end{center}

The {\itshape Pre-approach} action involves samplers for the constraints $Var_q^g$ and {\itshape ArmMotion}.
The grasp configuration $x_q^g$ is sampled in $Var_q^g$, which determines the direction of the TCP and the method of detection.
For example, top detection corresponds to a different state compared to forward or backward detection.
$Var_q^g$ is an optional constraint, depending on the specific task, which can reduce the size of the sampling space.
Similarly, we combine the samplers of $Var_q^g$ and {\itshape ArmMotion} to create a new, integrated conditional sampler for the picking process:
\begin{center}
        $\psi_{Pick}= \langle(x_0^a,x_q^a,x_q^g),(u_t^a),\{TrajectoryValid\}$,SAMPLE-PICK$\rangle$
\end{center}
where $x_0^a$ is the initial state of the arm and $x_q^a$ is the sampled pre-approach pose,
SAMPLE-PICK contains the generating process of $x_q^a$ and $x_q^g$,
it also calculates a collision-free path $u_t^a$ between the start state $x_0^a$ and the target state $x_q^a$.

Each atomic action in our framework has its planner during execution;
these planners function like $f(\vec{z}_I )$ of conditional samplers to generate the control parameters $\vec{u}_t$.
The grasp planner propagates the initial arm state backward to the intermediate states ($x_a^2$, $x_a^3$) and to the final grasp state $x_a^4$ (\Cref{fig_constraint_network_eg}).
$x_a^2$ is sampled by the conditional sampler $\psi_{Pick}$,
for example, the sampler can select top-grasp or forward-grasp to determine the TCP pose based on the grasping configurations.
$x_a^3$ and $x_a^4$ depend on the observing results.

Once all the free parameters of the action {\itshape Move-and-Pick} have been determined, we can formulate the BT to execute the action.

\subsection{Conditional Action Node and Conditional Switch Node}\label{section: CS Node and CA Node}

We design the CANode and CSNode to ensure the robust execution of factorable actions.
A CANode is an action node in BTs.
In addition to checking the preconditions of the action node, it also checks its corresponding boolean state at the beginning of the action.
If the boolean state is true, the CANode returns {\itshape Success} immediately;
if not, the CANode continues with the main part of the action like a normal action node and sets the boolean state to true upon successful completion.
CANodes run sequentially and achieve different effects by controlling the node state and associated free parameters.
The control module for CANodes is the CSNode.
The CSNode shares a readable and writable status table with its CANodes, with the values in this status table corresponding to the boolean states of each CANode in order.
In addition, the CSNode contains samplers for the free parameters of CANodes.
The CSNode is based on the condition node of BTs.
The CSNode functions as both an integrated switch that controls whether the CANodes are activated and as a controller that sets the free parameters of the CANodes.
Since the samplers for different factorable actions vary, the CSNode is based on a condition node rather than a control flow node.

The execution framework for factorable actions consists of modular subtrees of BTs with CANodes and CSNodes, which we refer to as {\itshape Conditional Subtrees} (CSubBTs).
The {\itshape blackboard} in BTs is used to update and deliver data streams.
An example structure of our proposed framework is illustrated in \Cref{fig_framework}.
Each atomic action within the factorable action is implemented through a CANode.
The root node for CSubBTs is a {\itshape Sequence} control node.
In the first ticking loop, the states of all CANodes are set to false.
If no anomalies occur, the CANodes run sequentially.
We add a {\itshape Decorator} node to ensure the CSubBT executes until it returns {\itshape Success}, or returns {\itshape Failure} if a conflicting constraint is detected.
Consequently, the CSNode can switch the boolean states of CANodes to control whether the atomic actions need to be re-executed when anomalies occur.

\begin{figure*}[!htp]
        \centering
        \begin{tabular}{@{\extracolsep{\fill}}c@{}c@{\extracolsep{\fill}}}
                \includegraphics[width=0.5\linewidth]{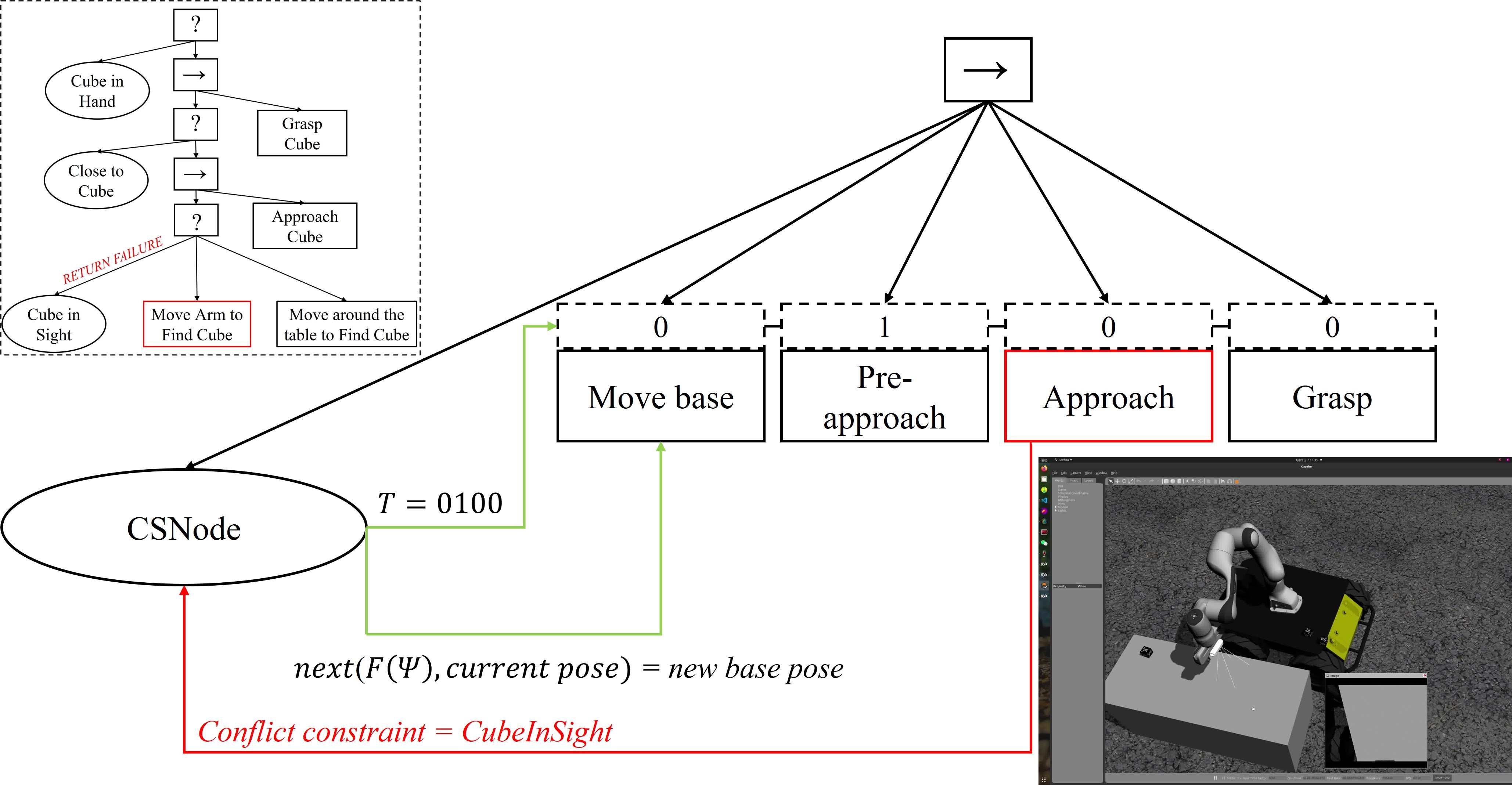} &
                \includegraphics[width=0.5\linewidth]{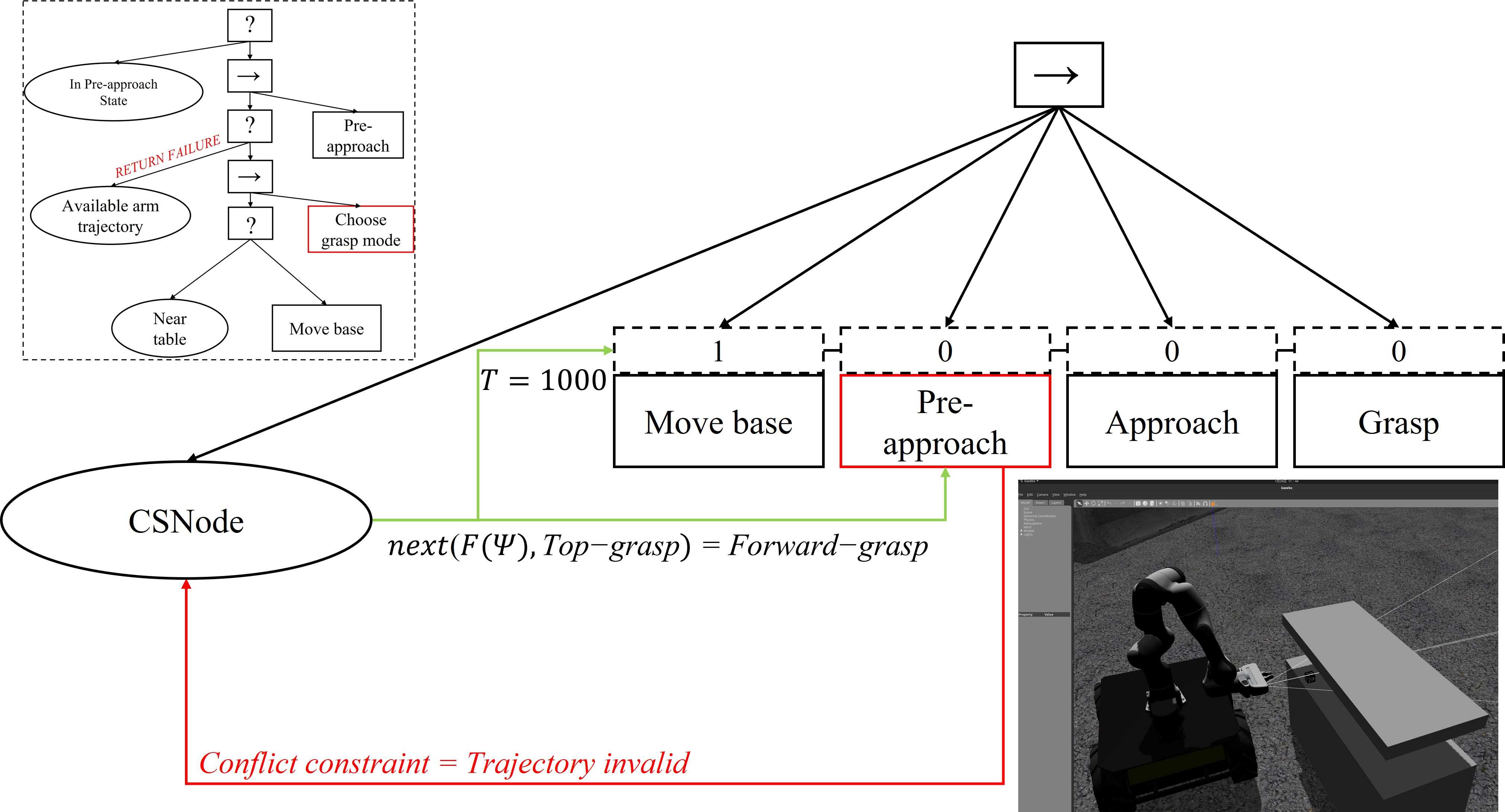}       \\
                (a)                                                 & (b) \\
        \end{tabular}
        \caption{The examples demonstrate how CSubBT handles two common exceptions.
                The objective of the mobile manipulator is to collect all the cubes on the desk.
                Error information and newly adjusted input data are transmitted via the {\itshape blackboard}.
                In example (a), when the robot fails to find a cube by swinging its arm in Cartesian space,
                the CSNode updates the status table's value (0100) to identify a new base pose that can address the situation.
                In example (b), the robot lacks a valid trajectory when performing the {\itshape Pre-approach} action with the top grasp mode.
                The CSNode generates an alternative grasp mode (e.g., forward grasp) to handle this exception.
                The dashed box in the figure illustrates the equivalent BT of CSubBT for handling the current exception, 
                designed based on traditional BT design principles (PPA and Backchaining).
                BTs that are fully equivalent to the CSubBT tend to have a more complex structure.}
        \label{fig_example}

\end{figure*}

Note that the structure presented in \Cref{fig_framework} is a simplified representation of CSubBT.
We can also add additional nodes to the CSubBT for other purposes.
We need to ensure that the CANodes corresponding to the atomic actions are ticked in sequence,
and that this sequence aligns with the CSubBT's order of state values in the shared status table.

The execution process of CSubBT involves finding feasible solutions through the sequence of conditional samplers $F(\vec{\psi } )$ for factored transition systems.
We sample the intermediate states in the constraint space and use the initial state as input parameters of $F(\vec{\psi } )$.
The motion planners function as components of $F(\vec{\psi } )$ to generate the set of control parameters $\vec{u}_t$.
As mentioned in \Cref{background}, CSubBT is robustly feasible if $F(\vec{\psi } )\cap \vec{\mathcal{C} }\neq \emptyset$.
By leveraging the {\itshape Move-and-Pick} action, we demonstrate how our framework can successfully complete the task even when encountering an exception.

\subsection{Framework Example}

The initial sample values of conditional samplers, as well as the BTs file, are manually defined in this paper;
they can be obtained through the high-level planner in practice.
If all the constraints in \Cref{fig_constraint_network_eg} are satisfied,
the {\itshape Move-and-Pick} CSubBT will execute sequentially and return {\itshape Success}.
The CANodes of the CSubBT are the atomic actions mentioned in \Cref{The Move-and-Pick Process} respectively.
\Cref{alg1} outlines the parameter updating process of CSubBT when an anomaly is detected.
The {\itshape NearGrasp} and {\itshape CloseCube} constraints serve as both preconditions for the current atomic actions and postconditions for the previous atomic actions.
We call these constraints as {\itshape logistic constraints}.
Similar to manual exception handling strategies, CSubBT can re-execute the previous CANode to address these constraints when they are not met (\Cref{alg1} Line \ref{repeatPreviousAction}).

\begin{algorithm}
        \caption{Parameters updating process}
        \label{alg1}
        \begin{algorithmic}[1]
            \REQUIRE The subtree ${\mathcal{T}_n}$, the set of {\itshape logistic constraints} $\vec{\mathcal{C}_d}$,
            the sequence of conditional samplers $\Psi(\mathcal{S}) = (\psi _1,...,\psi _n)$,
            the conflicted constraint $c$ and failed atomic action index t.
            \ENSURE the input parameters $params$ and the action status table $T$
            \IF{$c \in \vec{\mathcal{C}_d}$}
            \STATE ${\mathcal{T}_n}$.resetNodeStatus($T,t-1$)\label{repeatPreviousAction}
            \ELSE
            \STATE ${\bar{x}}_I = \emptyset $
            \STATE $\bar{\psi } = {\mathcal{T}_n}$.getCorrespondingSamplers$(\Psi(\mathcal{S}), c)$
            \FOR{each $\psi$ in $\bar{\psi }$}
                \STATE ${\bar{x}}_I = \mathbf{next}(\psi, current\_states)$\label{exploreSolution}
                \IF{ ${\bar{x}}_I \neq  \emptyset $}
                \STATE $\mathbf{break}$
                \ENDIF
            \ENDFOR
            \IF{ ${\bar{x}}_I \equiv   \emptyset $}
            \STATE $\mathbf{return}$ $\mathbf{Failure}$ \COMMENT{All parameters have been attempted}
            \ENDIF

            \STATE $T, params \gets {\mathcal{T}_n}$.updateBlackboardValues(${\bar{x}}_I$)\label{alg2updateBlackboardValues}
            \ENDIF
    
        \end{algorithmic}
    \end{algorithm}

If the {\itshape TrajectoryValid} and {\itshape PathValid} constraints are not met,
the corresponding conditional samplers $\psi_{Move}$ and $\psi_{Pick}$ will sample the next values as input parameters to the CANodes for a new attempt (\Cref{alg1} Line \ref{exploreSolution}).
The next sampled parameter is typically the one closest to the current robot state.
We can manually set re-execution policies for CANodes if the {\itshape CubeInSight} constraint is not satisfied.
For example, waving the arm in Cartesian space or moving the base pose are suitable choices.
Thus the {\itshape CubeInSight} constraint is also related to the conditional samplers,
we can try new values from the conditional samplers to fulfill the constraint.
The size of the sampling space can be determined by combining relevant geometric parameters,
such as the size of the robot, the visual field, the size of the desk, etc.
Integrating the execution framework with conditional samplers provides the method with exploration capabilities in the solution space when encountering anomalies.

\Cref{fig_example} illustrates the process by which CSubBT updates its parameters and manages the execution of atomic nodes when faced with two common anomalies.
The CSNode uniformly handles failure information, acting as an implicit 'switch-case' mechanism.
Similar to the exception-handling method of a deliberately designed BT (the dashed box in \Cref{fig_example}),
CSubBT provides corresponding solutions for unsatisfied constraints when anomalies occur.
The key difference is that CSubBT directly leverages the common reusability of atomic action nodes and factored transition systems to solve problems.
Although the centralized CSNode compromises the readability and intuitiveness of the control logic,
the structure of the BT becomes much simpler, allowing for a more robust execution of actions.
Furthermore, CSubBT remains modular and scalable to some extent, and it is more intuitive as a downstream executor for task implementation.

\begin{figure*}[!htp]
        \centering

        \begin{tabular}{@{\extracolsep{\fill}}c@{}c@{}c@{}@{\extracolsep{\fill}}}
                \includegraphics[width=0.31\linewidth]{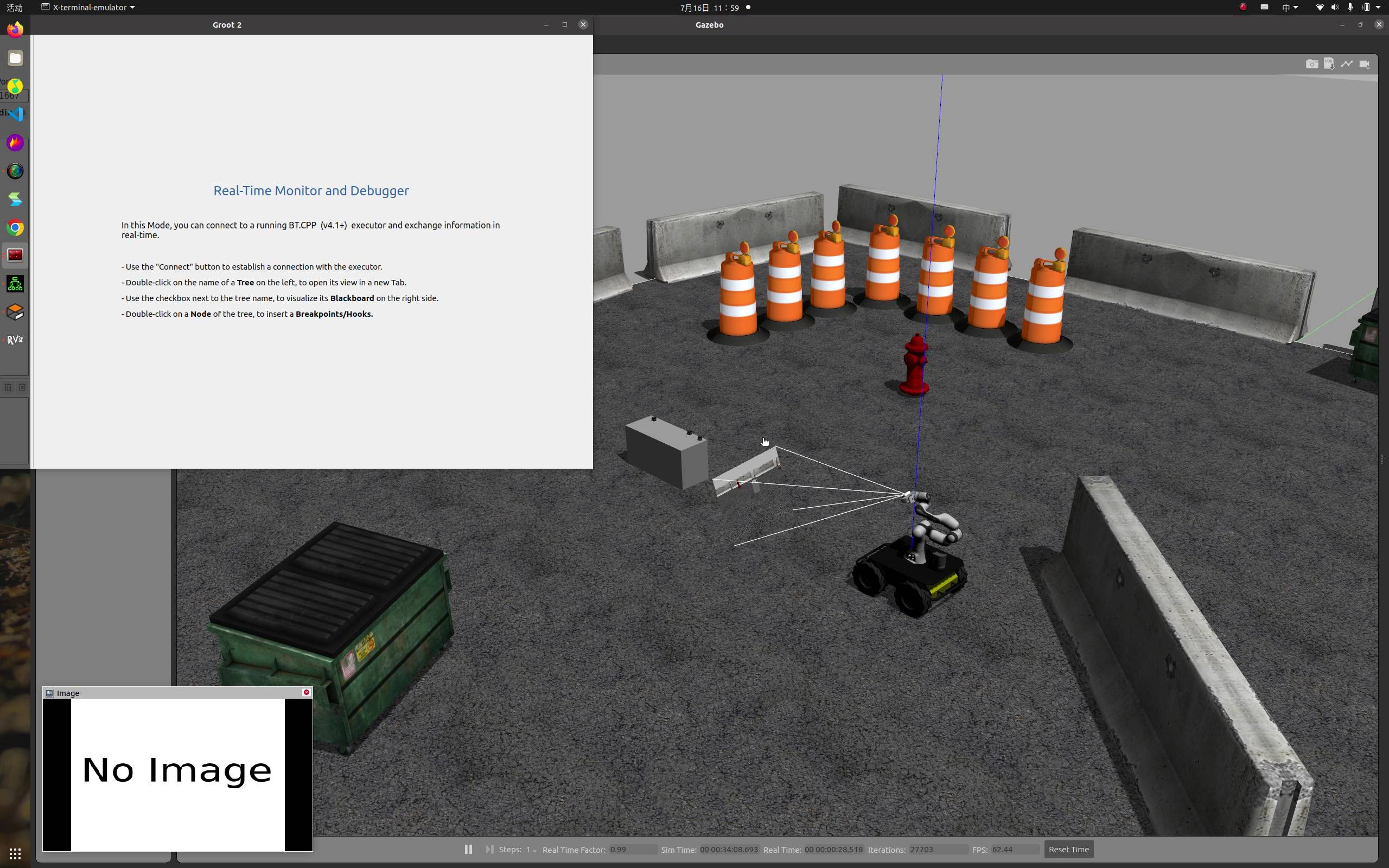} \hspace{.105in} &
                \includegraphics[width=0.31\linewidth]{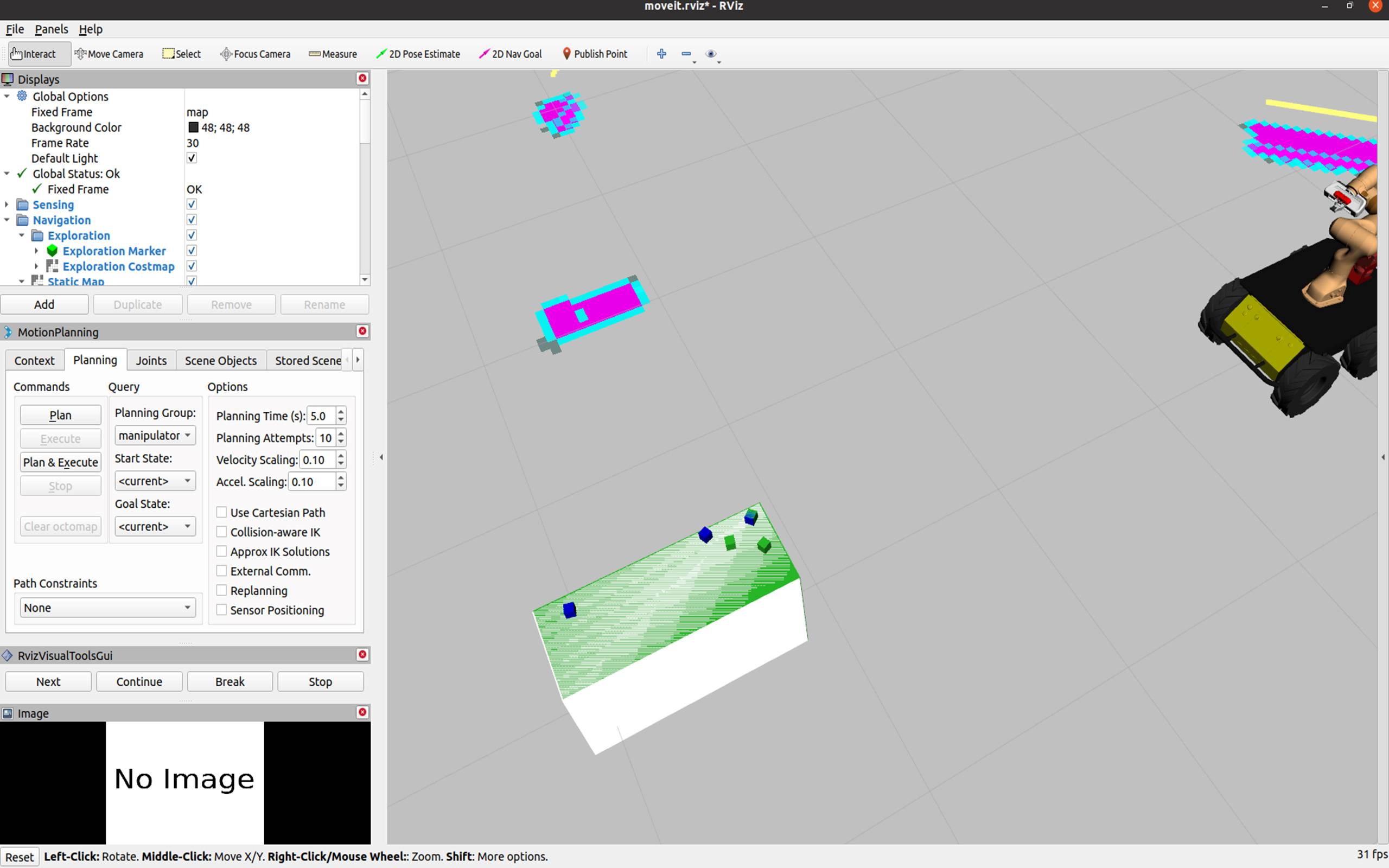} \hspace{.105in} &
                \includegraphics[width=0.31\linewidth]{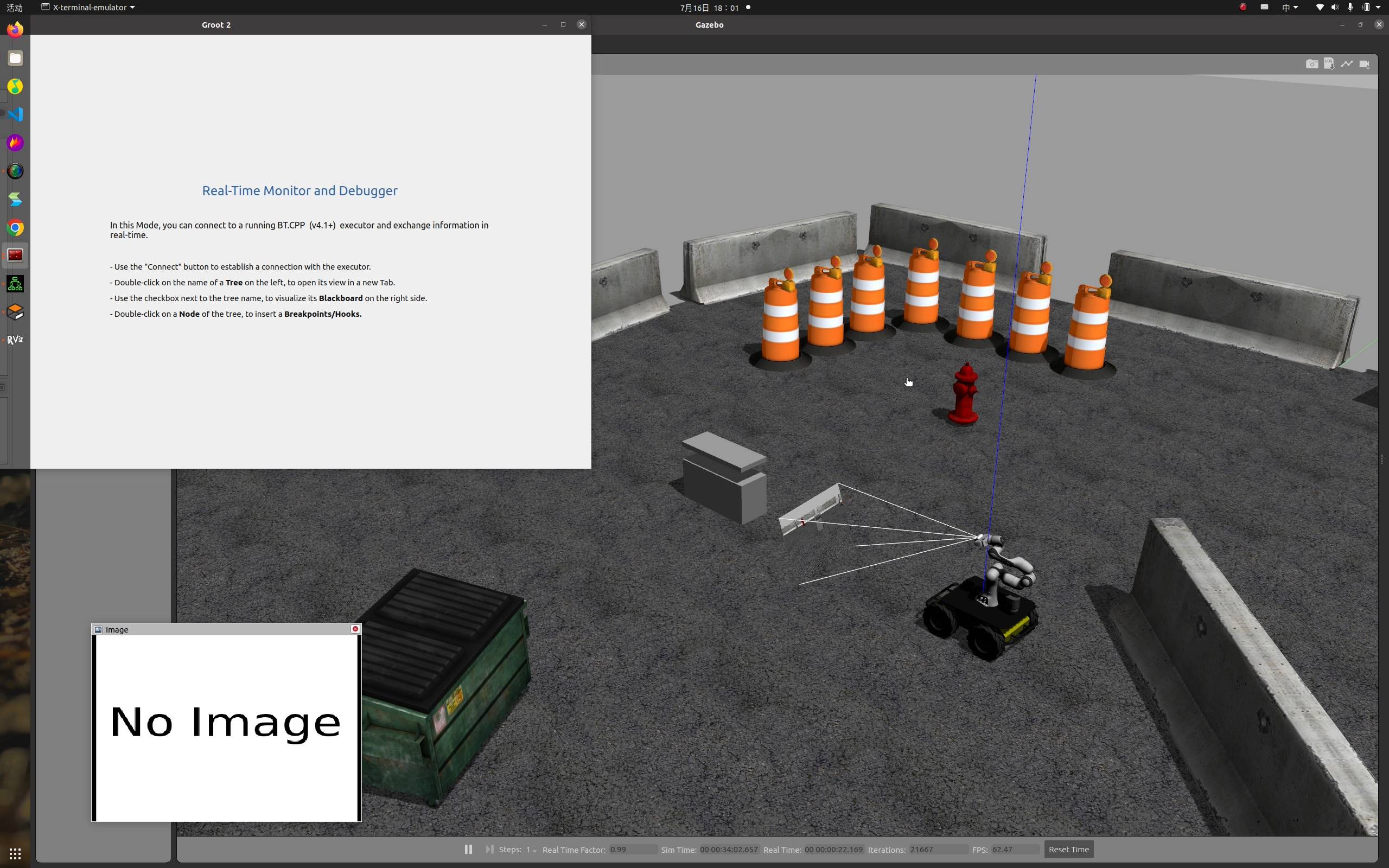}                             \\
                (a)                                                                          & (b) & (c) \\
        \end{tabular}

        \begin{tabular}{@{\extracolsep{\fill}}c@{}c@{}c@{}c@{}@{\extracolsep{\fill}}}
                \includegraphics[width=0.231\linewidth]{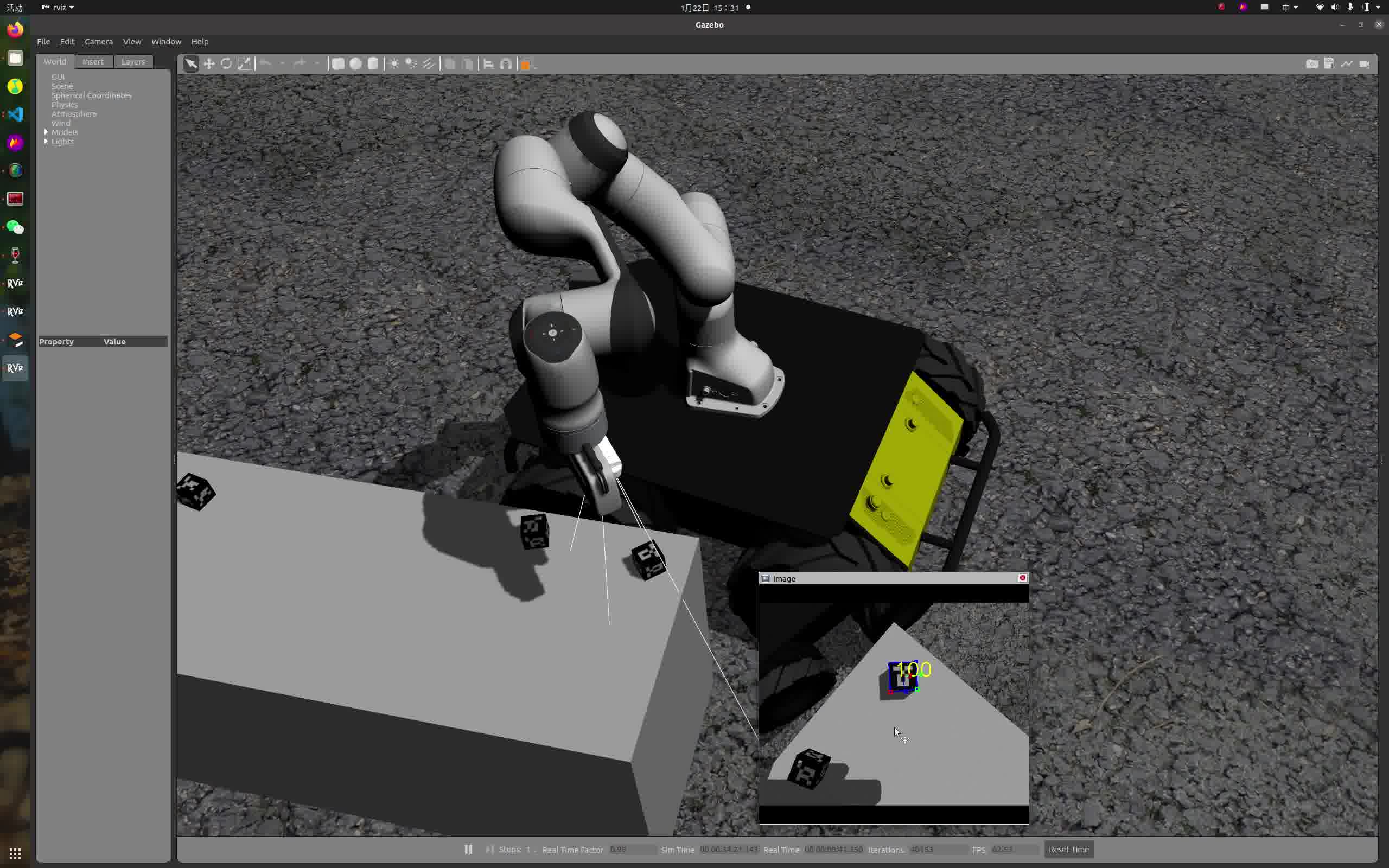} \hspace{.07in} &
                \includegraphics[width=0.231\linewidth]{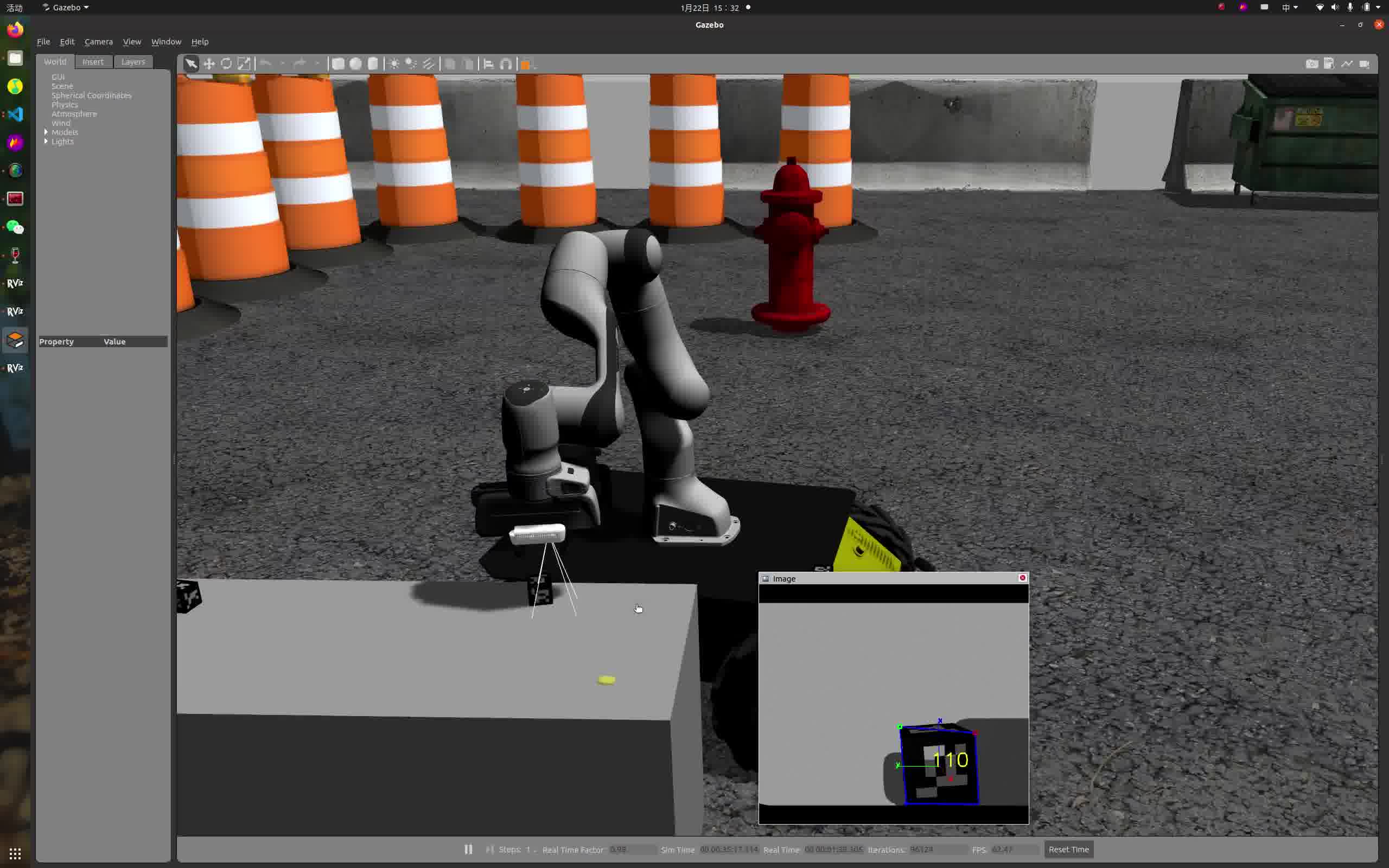} \hspace{.07in} &
                \includegraphics[width=0.231\linewidth]{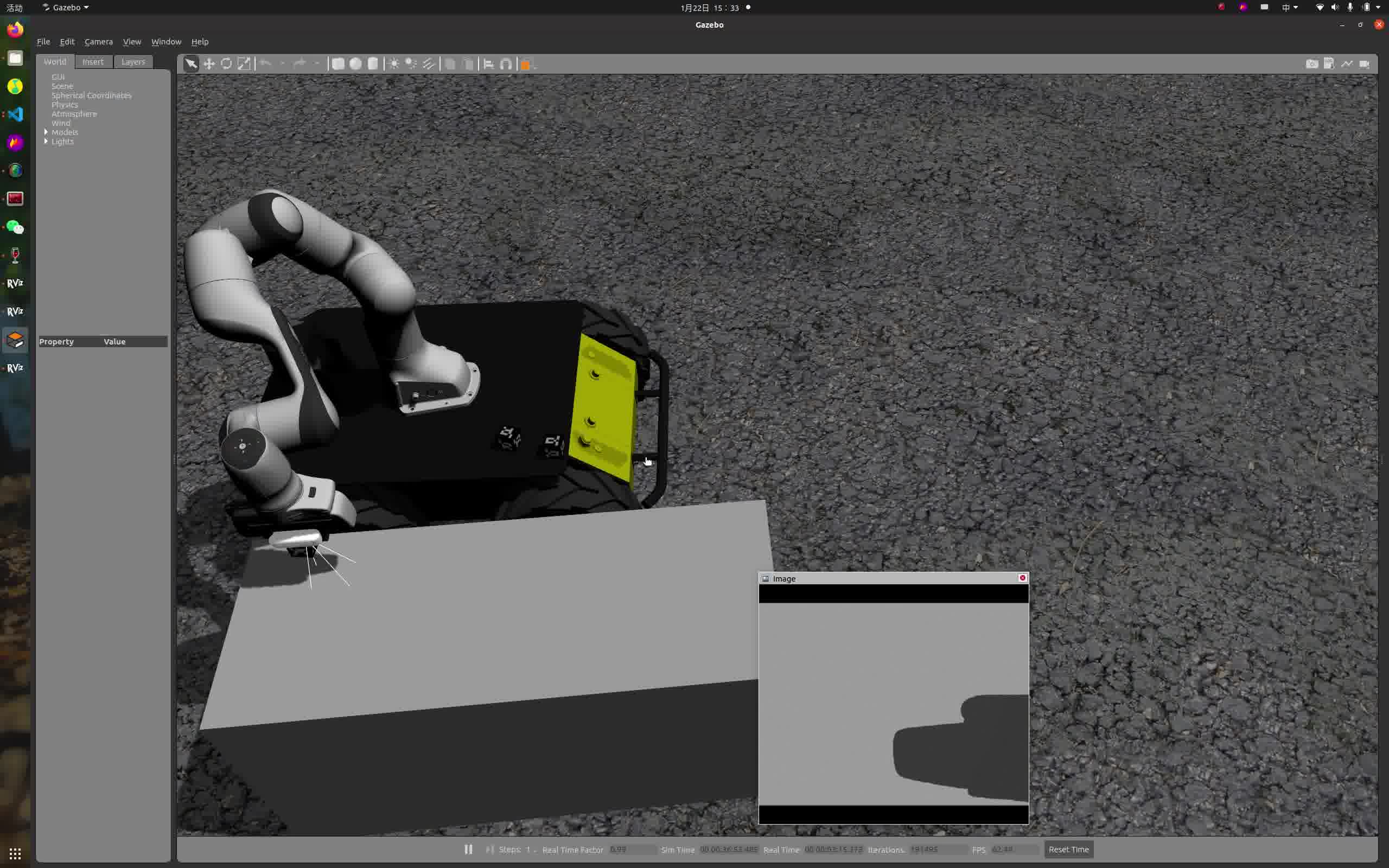} \hspace{.07in} &
                \includegraphics[width=0.231\linewidth]{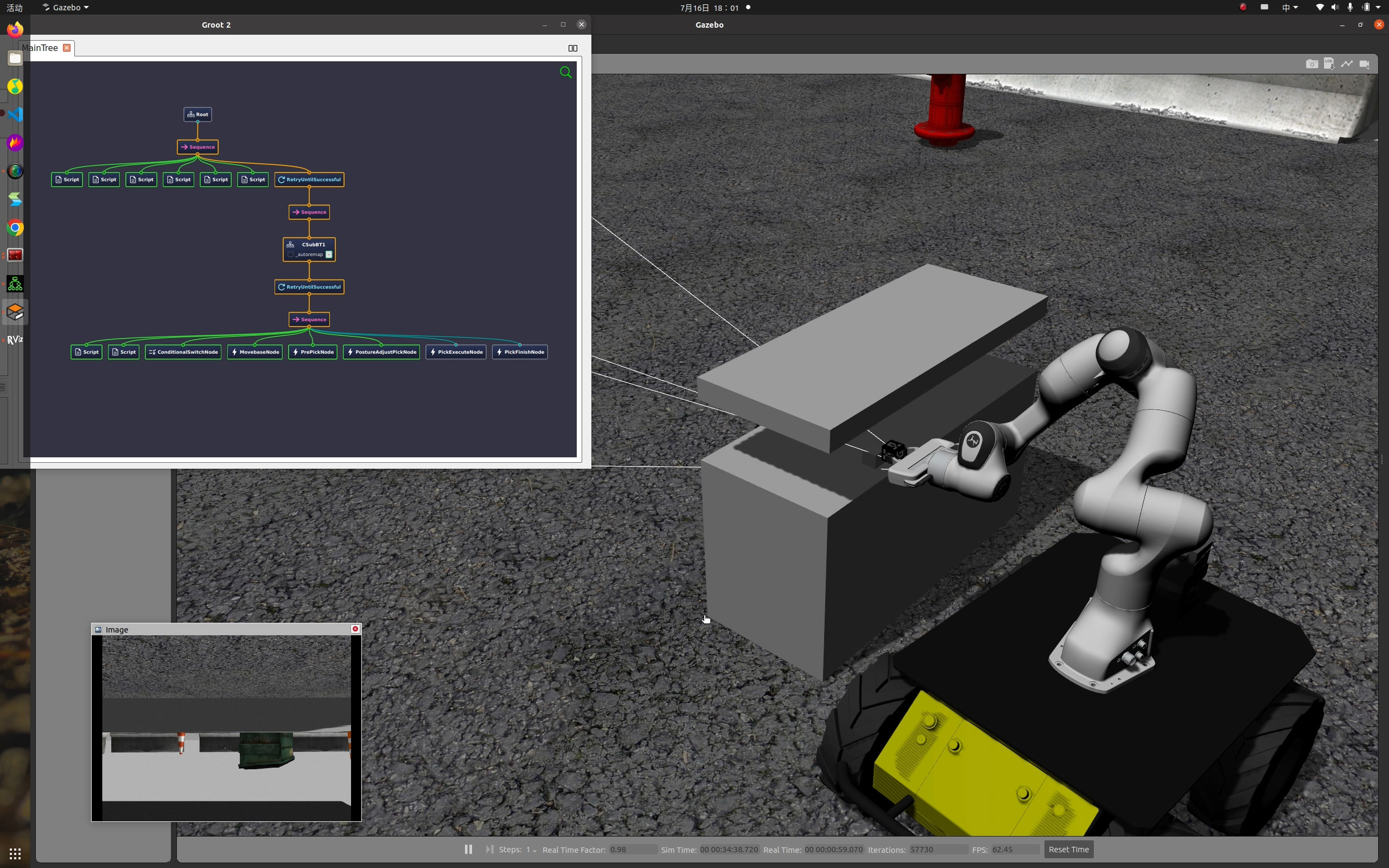}                                  \\
                (d)                                                                          & (e) & (f) & (g) \\
        \end{tabular}
        \caption{Gazebo simulation experiment scenarios using a mobile manipulation platform are presented.
                (a) shows the initial state of the classic mobile grasping experiment.
                (b) depicts the three cubes in the task, each with varying degrees of perceptual bias.
                The small green cube indicates the robot's expected position, while the small blue square represents the actual position.
                (c) illustrates the initial state of the capture occlusion experiment.
                (d)-(g) show parts of the grasping process for the cubes, the scripts nodes in the Groot2 monitor window of figure (g) are the initial parameters in the blackboard.
        }
        \label{fig_experiment}
\end{figure*}

\section{Experiments}

This section demonstrates the effectiveness of our execution framework through both simulation experiments\footnote{[Online]. Available: https://youtu.be/227PGCuv7dw} and real-world experiments\footnote{[Online]. Available: https://youtu.be/XeKP3zhcMQ0}.
We conducted robustness and reactivity experiments in Gazebo environment using the MoveIt! grasp planner.
The mobile manipulator platform comprises a Husky mobile base, a 7-DoF Franka Emika Panda manipulator, and a standard gripper equipped with a Realsense camera.
BTs are implemented using the BehaviorTree.CPP\footnote{[Online]. Available: https://www.behaviortree.dev/} library of version 4.0.2,
and the Groot2 graphical interface monitors the execution state of BTs.

\subsection{Robustness Experiments}

We conducted four cube-grasping scenarios (\Cref{fig_experiment}) to evaluate our system's robustness in long-horizon mobile manipulation tasks.
The pre-generated plan may include target objects with varying degrees of perceptual bias, and dynamic environmental changes may occur during execution.
In the first scenario, the perceptual bias of the cube exemplifies the sensor deviations commonly occurring during mobile manipulation tasks.
These deviations are typically minor but can lead to the failure of pre-generated plans.
In the second scenario, the cube has a moderate perceptual bias, requiring the robot to reposition the target by moving its arm.
In the third scenario, the cube exhibits a significant perceptual bias, necessitating both arm and base movements to locate the target.
In the fourth scenario, the cube is partially obscured by an obstacle, requiring a re-direction of the grasp.
In several experiments, in addition to the designed anomalies, some unintended anomalies were encountered.
These include missed visual detections and sliding cubes. Nevertheless, our framework adapts to such unintended anomalies and achieves the desired outcome.

\begin{table}[h]
        \caption{ROBUSTNESS EXPERIMENTS}
        \centering
        \label{table_exp}
        \resizebox{\linewidth}{!}{
                \begin{tabular}{ccc|c}
                        \toprule 
                        Scenarios & Trials & Success & Expected Strategies               \\ 
                        \midrule 
                        1         & 10     & 10      & Relocate                          \\
                        2         & 10     & 9       & Relocate \& Wave arm              \\
                        3         & 10     & 8       & Relocate \& Wave arm \& Move base \\
                        4         & 10     & 10      & Redirect grasping  \&  Relocate   \\
                        \midrule
                        Total     & 40     & 92.5\%                                      \\
                        \bottomrule 
                \end{tabular}
        }
\end{table}

Additionally, we conducted independent grasping experiments to gather statistical data.
In these experiments, each cube was randomly placed on one side of the table.
\Cref{table_exp} presents the results of 40 grasping experiments, showing an overall success rate of 92.5\%.
The remaining 7.5\% failure rate is attributed to cube detection accuracy issues.
Due to missed detections, the robot adjusted its position after swinging its arm, which resulted in a position even further from the target, causing a failed grasp.
Our framework partially mitigates the reduction in detection rates caused by sunlight occlusion by waving the arm.
Theoretically, finer adjustment parameters could be set to avoid this situation.
Nonetheless, this result demonstrates the robustness of our execution framework.

\subsection{Reactivity Experiments}

Since CSubBT is a reactive actuator based on the TAMP method from \cite{b5,b7}, it remains fundamentally a reactive planner. 
We compare its runtime with that of classical reactive TAMP planners across several grasping tasks. 
For the comparison, we select pddlstream \cite{b7} with the Adaptive algorithm as the first reactive TAMP planner (Rep-pddlstream), 
which performs replanning when anomalies occur. 
Additionally, we include RH-TAMP \cite{b8}, a hierarchical reactive TAMP planner, 
and use the same FastDownward configurations for both RH-TAMP and Rep-pddlstream. 
To further evaluate reactivity, we manually developed a traditional BT (TradiBT) based on classical BT design principles to handle anomalies during the experiments.

\begin{figure}[!h]
        \centerline{\includegraphics[width=0.62\linewidth]{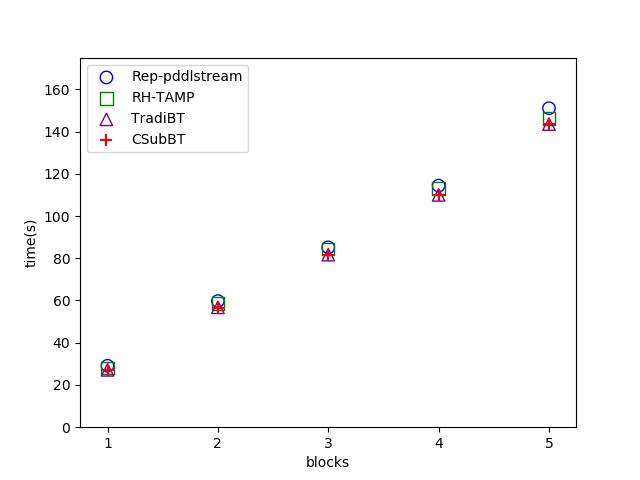}}
        \caption{Reactivity experiments: average runtime of the methods over 10 trials per problem size.
        }
        \label{fig_reactivity_exp}
\end{figure}

In these reactivity experiments, the robot is tasked with collecting varying numbers of targets from the table. 
We assume that the robot does not need to move, but that perceptual biases affect the targets during execution. 
As a result, the reactive planners must perform replanning during execution. 
To avoid task failures, we incorporate the same adjustment strategy used in CSubBT into the other methods. 
All trials were conducted on a 2.3 GHz Intel Core i7 processor, and the results are shown in \Cref{fig_reactivity_exp}.

The results reveal minimal differences in runtime among the four methods when grasping a single target, 
as replanning does not incur significant time overhead. 
However, as the number of grasping tasks increases, 
CSubBT demonstrates superior runtime efficiency compared to the other reactive planners. 
There is little difference in operational efficiency between CSubBT and traditional reactive behavior tree designs. 
Although CSubBT sacrifices some readability, it proves to be a more suitable modular execution mechanism for planners.

\subsection{Real-world Validation}

\begin{figure}[!h]
        \centerline{\includegraphics[width=0.62\linewidth]{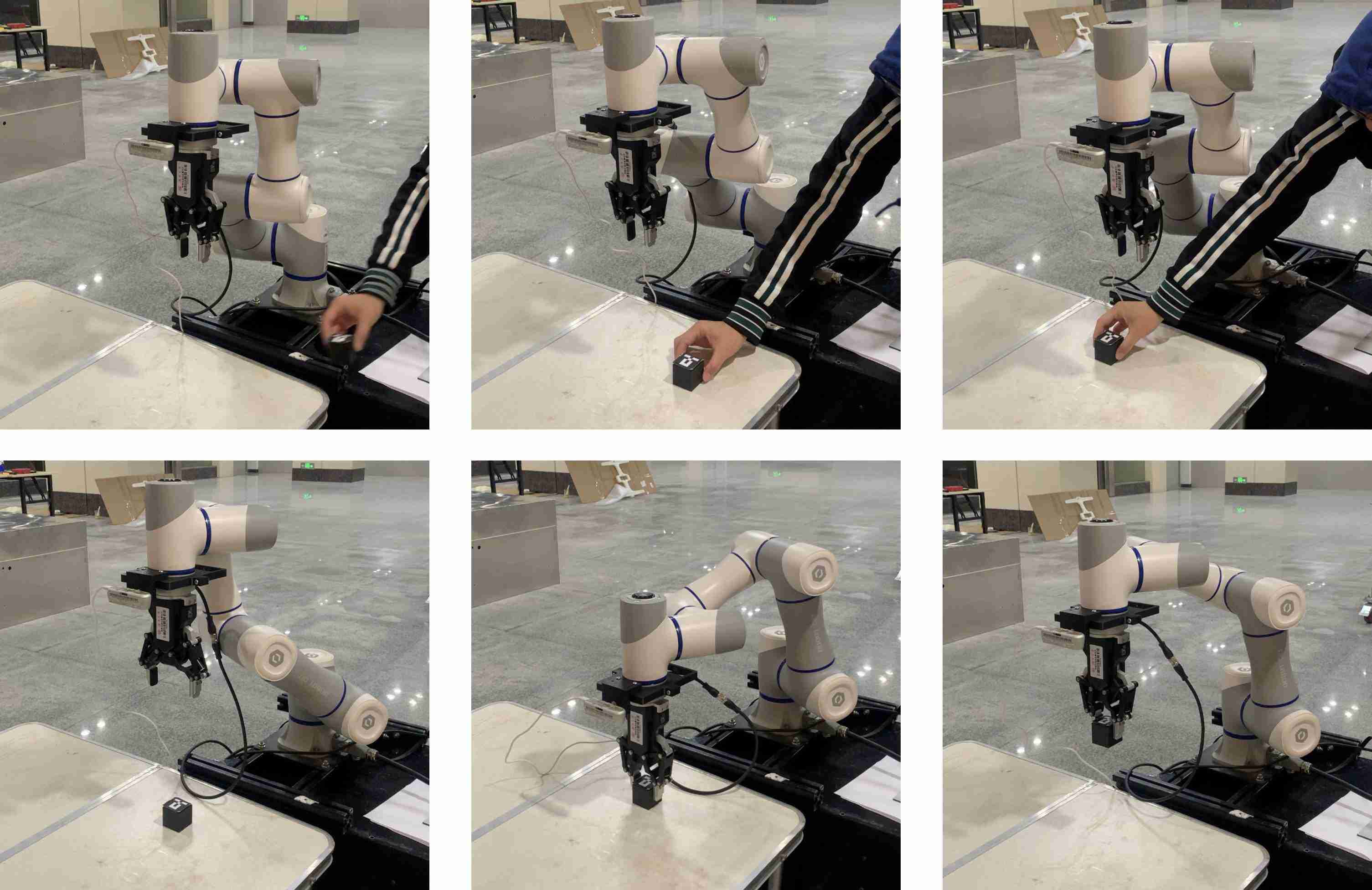}}
        \caption{Dynamic grasping experiments in a real unstructured environment.
                In our execution system, the robotic arm first attempts to locate the cube,
                then after human interference with the cube position the system self-adjusts and finally completes the grasping task.
        }
        \label{fig_real_world_exp}
\end{figure}

In the real-world scenario illustrated in \Cref{fig_real_world_exp}, we deployed a DOBOT CR3 6-axis robotic arm and a DH-ROBOTICS AG-95 hand gripper on a mobile platform.
During the experiments, the position of the object to be grasped by the arm was deliberately perturbed.
If the object was not detected, the arm moved within Cartesian space to locate it.
If the object's pose changed dynamically during the grasping process, the arm adjusted accordingly.
These experiments demonstrated that our system can successfully perform tasks even when encountering human-induced dynamic disturbances within tolerance limits.

\section{Conclusion And Limitation}

In this paper, we introduce CSubBT, a BT-based execution framework for mobile manipulation tasks.
The framework models anomalies as constraint violations and efficiently utilizes constraint-related samplers, 
commonly employed in higher-level planners, to address these violations.
CSubBT decomposes symbolic actions into CANodes and handles abnormal situations during execution by 
integrating these samplers within CSNodes to dynamically adjust execution parameters and control the execution of CANodes.
Experimental results validate the robustness and enhanced reactivity of CSubBT in task execution.

However, CSubBT has certain limitations.
The decomposition of symbolic actions relies on the engineering implementation of an action, 
and the size of the corresponding constraint space is dependent on the number of atomic actions involved.
Additionally, CSubBT currently requires manual configuration of sampling parameters
and lacks the capability to autonomously adjust these settings based on environmental conditions during execution.
While CSubBT can explore the constraint space of actions, it is unable to handle anomalies that lie outside this space. 
Nevertheless, by thoroughly exploring the constraint space, the execution system can offer more accurate feedback on anomalies.

Recent advancements in Large Language Models (LLMs) have shown significant potential in the field of planning. 
Learning algorithms could help address the limitations of static sampling configurations. 
Future work will focus on integrating high-level planners with CSubBT and 
leveraging the general reasoning capabilities of Large Language Models (LLMs) 
to develop a more comprehensive and adaptive planning and execution system.

\bibliographystyle{unsrt}  
\bibliography{references}

\end{document}